\documentclass[runningheads]{llncs}
\usepackage{amsmath}
\usepackage{graphicx}
\usepackage[colorlinks=true]{hyperref}
\usepackage{colortbl}
\usepackage{multirow}
\usepackage{booktabs}
\usepackage{array}
\usepackage{xcolor} 
\usepackage{newtxtext}
\usepackage{amssymb}
\usepackage{microtype}
\usepackage[T1]{fontenc}
\usepackage{graphicx}
\begin{document}
\title{Retinexmamba: Retinex-based Mamba for Low-light Image Enhancement}
\titlerunning{RetinexMamba}
%
\author{Jiesong Bai\inst{1,*} \and
Yuhao Yin\inst{1,*} \and
Qiyuan He\inst{1} \and
Yuanxian Li\inst{2} \and
Xiaofeng Zhang\inst{3,\dagger}
}
\authorrunning{Bai et al.}

%
\institute{Shanghai University \and 
Yunnan University \and
Shanghai Jiao Tong University
\\
{* Equal contribution.   }
{                † Corresponding author}
}

\maketitle              

\begin{abstract}
\textit{In the field of low-light image enhancement, both traditional Retinex methods and advanced deep learning techniques such as Retinexformer have shown distinct advantages and limitations. Traditional Retinex methods, designed to mimic the human eye's perception of brightness and color, decompose images into illumination and reflection components but struggle with noise management and detail preservation under low light conditions. Retinexformer enhances illumination estimation through traditional self-attention mechanisms, but faces challenges with insufficient interpretability and suboptimal enhancement effects. To overcome these limitations, this paper introduces the RetinexMamba architecture. RetinexMamba not only captures the physical intuitiveness of traditional Retinex methods but also integrates the deep learning framework of Retinexformer, leveraging the computational efficiency of State Space Models (SSMs) to enhance processing speed. This architecture features innovative illumination estimators and damage restorer mechanisms that maintain image quality during enhancement. Moreover, RetinexMamba replaces the IG-MSA (Illumination-Guided Multi-Head Attention) in Retinexformer with a Fused-Attention mechanism, improving the model's interpretability. Experimental evaluations on the LOL dataset show that RetinexMamba outperforms existing deep learning approaches based on Retinex theory in both quantitative and qualitative metrics, confirming its effectiveness and superiority in enhancing low-light images.Code is available at \url{https://github.com/YhuoyuH/RetinexMamba}}

\end{abstract}

\begin{figure}
    \centering
    \includegraphics[width=1\linewidth]{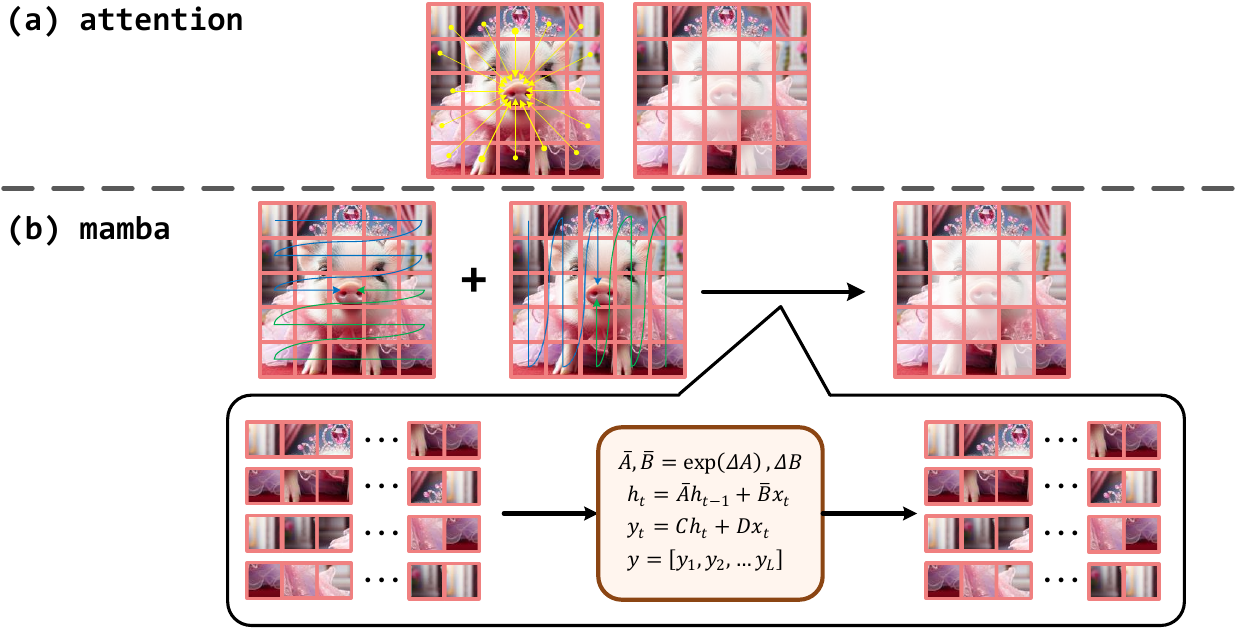}
    \caption{The image above shows a visual comparison between the attention mechanism and 2D selective scanning in Mamba. In Mamba's 2D selective scanning, scanning starts simultaneously from all sides of the image, whereas the attention mechanism calculates attention scores separately from the target view to the global view.}
    \label{fig:fig.5}
\end{figure}

\section{Introduction}
\hspace{1em}Low-light enhancement refers to the process of improving the quality and visual appearance of images captured under insufficient light or dim environmental conditions through image processing techniques. This field is of significant importance in computer vision and image processing because images under low-light conditions often suffer from issues such as dimness, blurriness, and unclear details, which affect the quality and usability of the images.

Traditional techniques such as histogram equalization\cite{ref1,ref2,ref3,ref4,ref5,ref7,ref9} and gamma correction\cite{ref6,ref8,ref10}, while foundational and important, often fall short in  dealing with complex lighting dynamics and maintaining the naturalness of enhanced images. Inspired by the human visual system, the Retinex theory\cite{ref56} establishes a conceptual framework for separating the illumination and reflection components of an image, paving the way for more sophisticated enhancement strategies capable of addressing diverse and challenging low-light environments.

Recent advancements in neural networks, particularly in the application of Convolutional Neural Networks (CNN) and Transformer models, have set new benchmarks in low-light image enhancement. CNNs\cite{ref21,ref24,ref25,ref26} exhibit strong capabilities in low-light image enhancement by effectively capturing spatial information and local features within images. However, CNNs may have limitations in modeling long-range dependencies, leading to insufficient consideration of overall image information and challenges in effectively addressing issues like noise amplification and detail preservation in low-light image processing. On the other hand, Transformer models\cite{ref29,ref36,ref62,ref66,ref68} achieve global perception and modeling of long-range dependencies through self-attention mechanisms, aiding in more accurately restoring details and structures in images during low-light image processing. However, Transformer models also face challenges such as high computational complexity and large parameter sizes, resulting in potential issues of slow inference speed and high resource consumption in practical applications.

In this context, our work introduces the \textit{RetinexMamba} architecture. Initially following the approach of Retinexformer, we divide the overall architecture into an illumination estimator and a damage repairer. The illumination estimator is used to initially light up the image, while the damage repairer eliminates amplified artifacts and noise generated during the illumination process, as well as color distortions and overexposure. To address the high computational complexity in Transformers and the insufficient interpretability of the attention mechanism in Retinexformer, the basic unit of our damage repairer is the Illumination Fusion State Space Model (IFSSM). This model utilizes 2D Selective Scanning (SS2D) as the backbone network to achieve linear computational efficiency, and employs Illumination Fusion Attention (IFA) to replace the Illumination-Guided Multi-head Self-Attention (IG-MSA) in Retinexformer to enhance the interpretability of the attention mechanism.

Through comprehensive experimental setups, we provide quantitative and qualitative results that demonstrate the superiority of our model on standard benchmarks such as the LOL dataset\cite{ref30,ref52}. According to the data in Tab.\ref{tab:mytable1}, using this approach, we have surpassed the SOTA of deep learning methods based on Retinex theory on the LOL dataset. Our contributions can be summarized as follows:
\begin{enumerate}
    \item Introducing Mamba for low-light enhancement for the first time, using SS2D to replace Transformers in capturing long-range dependencies.
    \item We proposed a fusion module that better implements the embedding of illumination features consistent with Retinex theory.
    \item Extensive quantitative and qualitative experiments prove that our method surpasses all previous deep learning methods based on Retinex theory.
\end{enumerate}

\section{Related Work}

\subsection{Low-light Image Enhancement}
\subsubsection{Distribution Mapping Method.}
In early research on low-light image enhancement, one of the most intuitive approaches is to map the distribution of the low-light inputs in a way that amplifies smaller values (which appear dark). Representative techniques of this approach include histogram equalization \cite{ref1,ref2,ref3,ref4,ref5,ref7,ref9} and S-curve-based methods, such as gamma correction \cite{ref6,ref8,ref10}. However, existing distribution mapping-based methods often suffer from color distortion and other artifacts that affect the visual quality of the enhancement results, due to the lack of recognition and utilization of semantic information during the distribution mapping process.
\subsubsection{Traditional Model Method.}
Retinex theory \cite{ref56} provides an intuitive physical explanation for the process of enhancing weakly lit images. This theory postulates that desired normal images (i.e., reflectance maps) can be obtained by removing the illumination component from low-light inputs. Jobson et al.\cite{ref11,ref12} conducted exploratory research based on the Retinex model. As research progressed, it became apparent that the key to achieving brightness enhancement using the Retinex approach is the estimation of the illumination layer \cite{ref13,ref14,ref15,ref16,ref17}. Such methods rely on manually crafted priors and often require careful parameter tuning. Inaccurate priors or regularization can result in artefacts and color bias in the enhanced images, demonstrating poor generalization capabilities and a time-consuming optimization process. Additionally, these studies often overlook the presence of noise, which can result in its retention or amplification in the enhanced images.
\subsubsection{Deep Learning Method.}
Deep learning-based methods for low-light image enhancement originated in 2017 \cite{ref20} and subsequently became the leading approach in the field.
Following the traditional Retinex theory \cite{ref56} as a cornerstone for model architecture, a series of works were designed along these lines \cite{ref33,ref32}. CNN \cite{ref21,ref24,ref25,ref26,ref28,ref34,ref61,ref63} has been widely used in low-light image enhancement. For example, Wei et al. \cite{ref22} and subsequent works \cite{ref30,ref23} combined Retinex decomposition with deep learning. However, these CNN-based methods had limitations in capturing long-range dependencies across different regions. Star \cite{ref29} introduced the transformer architecture to the field of low-light enhancement, addressing the issue of capturing long-range dependencies. Meanwhile, Retinexformer \cite{ref36} combined the Retinex theory with the design of a one-stage transformer, further refining and optimizing this approach. Nonetheless, transformer models pose significant computational burdens and complexity when dealing with long sequences due to their self-attention mechanisms.
\subsection{State Space models}
\hspace{1em}Recently, the state space models (SSMs) have been increasingly recognized as a promising direction in research. A structured state space sequence model, referred to as S4, was proposed in \cite{ref39} as a novel alternative to CNNs or Transformers for modeling long-range dependencies.Subsequent developments have seen various structured state space models emerge, featuring complex diagonals \cite{ref48,ref49}, multi-input multi-output support \cite{ref50}, diagonal decomposition and low-rank operations \cite{ref51} enhancing their expressive capabilities.Contemporary SSMs such as Mamba \cite{ref40} not only establish long-distance dependency relations but also demonstrate linear complexity with respect to input size. Models based on SSM architecture have garnered extensive research interest across diverse domains \cite{ref44}.The selective scanning mechanism introduced by Mamba \cite{ref40} matches the performance of prevalent foundational models in vision, such as \cite{ref53,ref54,ref55}.Vision Mamba\cite{ref41} suggests that pure SSM models can serve as a generic visual backbone. Empirical validation is provided by \cite{ref43,ref47}, which have shown promise in the task of medical image segmentation, while in low-level vision tasks, applications such as \cite{ref45,ref46} have demonstrated favorable outcomes.Inspired by this research, our work leverages Mamba's capability for linear analysis of long-distance sequences to process features fused with Retinex theory \cite{ref56}. The resulting enhancements to low-light images substantiate the potential of Mamba-based models in the domain of low-light image enhancement.
\begin{figure}
    \centering
    \includegraphics[width=0.986\linewidth]{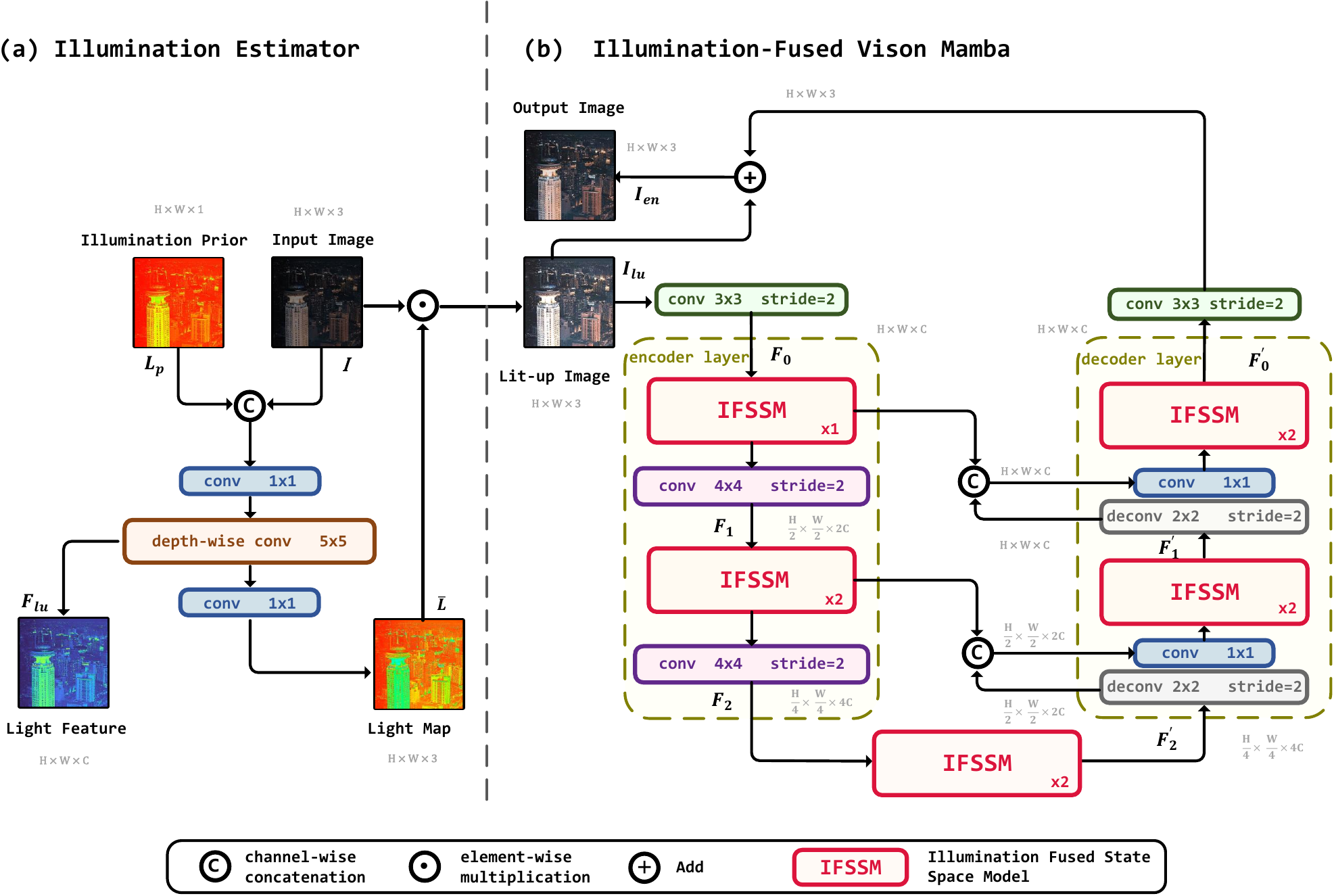}
    \caption{Our structural framework is displayed in the two main areas above. It comprises an illumination estimator (a) and a Damage Restorer based on the Illumination Fusion Visual Mamba (IFVM) (b).}
    \label{fig:enter-label}
\end{figure}

\section{Method}
\hspace{1em}Fig. \ref{fig:enter-label} depicts the comprehensive structure of our approach. As illustrated in Fig. \ref{fig:enter-label}, our RetinexMamba consists of an Illumination Estimator (a) and a Damage Restorer (b). The Illumination Estimator (IE) is influenced by the conventional Retinex theory. The design of the Damage Restorer is based on the Illumination Fusion Visual Mamba (IFVM). Displayed in Fig. \ref{fig:enter-label}(b), the core component of IFVM is the Illumination Fusion State Space Model (IFSSM), featuring a Layer Normalization (LN), an Illumination-Fused Attention Mechanism (IFA), a 2D Selective Scan(SS2D), and a Feed-Forward Network (FFN). The specifics of the IFA are detailed in Figure 2(c).

\subsection{Retinex-based Framework}
\hspace{1em}The traditional Retinex image enhancement algorithm simulates human visual perception of brightness and color. It decomposes image \({I} \in \mathbb{R}^{H \times W \times 3}\) into the illumination component \({L} \in \mathbb{R}^{H \times W}\) and the reflection component \({R} \in \mathbb{R}^{H \times W \times 3}\). This conclusion can be expressed by the following formula:
\begin{equation}
I = R \circ L
\end{equation}
where \(\circ\) denotes element-wise multiplication. The reflection component \(R\) is determined by the intrinsic properties of the object, while the illumination component \(L\) represents the lighting conditions. However, under this formulaic expression, the traditional Retinex algorithm does not account for the noise and artifacts produced by unbalanced light distribution or dark scenes in low-light conditions, and this loss of quality is further amplified with the enhancement of the image. Therefore, inspired by the Retinex algorithm, we have adopted the perturbation modeling proposed by \cite{ref36}, which introduces perturbation terms for the illumination and reflection components \(\tilde{L}\) and \(\tilde{R}\) in the original formula, as shown in the following equation:
\begin{align}
I &= (R + \tilde{R}) \circ (L + \tilde{L}) \\
  &= R \circ L + R \circ \tilde{L} + \tilde{R} \circ L + \tilde{R} \circ \tilde{L},
\end{align}
where \( \tilde{R} \in \mathbb{R}^{H \times W \times 3}\) and \(\tilde{L} \in \mathbb{R}^{H \times W }\)  represent the perturbation terms for the reflection and illumination components respectively. After simplification, we can express the illuminated image \( I_{\text{lu}} \) as follows:
\begin{equation}
I_{\text{lu}} = I \circ \bar{L} = R + C,
\end{equation}
where \( \bar{L} \) represents the illumination mapping, which is obtained through convolution for feature extraction, and \( {C} \in \mathbb{R}^{H \times W \times 3}\) denotes all the losses previously mentioned. Hence, our RetinexMamba can be expressed as:
\begin{align}
(I_{\text{lu}}, F_{\text{lu}}) &= IE(I, L_{\text{p}}), \\
I_{\text{en}} &= IFVM(I_{\text{lu}}, F_{\text{lu}}), \\
RM &= IE(I_{\text{lu}}, L_{\text{p}}) + IFVM(I_{\text{lu}}, F_{\text{lu}}),
\end{align}
where IE represents the Illumination Estimator, and IFVM denotes the Damage Restorer. IE takes the low-light image \( I \) and the illumination prior \( L_p \in \mathbb{R}^{H \times W}\) as input, outputting the illuminated image \( I_{lu} \in \mathbb{R}^{H \times W \times 3}\) and the illumination feature map \( F_{lu} \in \mathbb{R}^{H \times W \times C}\). The \( L_p \) is obtained by calculating the average value of each channel of the image, used to assess the overall brightness or illumination level of the image; hence we use \( L_p \) as an illumination prior to provide lighting information for the image. Then, these two results are fed into the Damage Restorer (IFVM) to mend the quality loss amplified during the illumination of the image, and finally generate the repaired result \( I_{en} \in \mathbb{R}^{H \times W \times 3}\).
\subsubsection{Illumination Estimator.}
The structure of the Illumination Estimator(IE) is shown in Fig. \ref{fig:enter-label}(a). We merge the low-light original image \( I \) with the illumination prior \( L_p \) obtained through calculation, and increase the channel dimension to serve as input. This is followed by three convolutions to extract features. The first \( conv \) \( 1\times1 \) merges the previously combined input, that is, to apply the illumination prior to the fusion into the low-light image. The second depthwise separable \( conv \) \( 5\times5 \) upsamples the input, further extracting features to generate the illumination feature map \( F_{\text{lu}} \), where the feature dimension \( n_{\text{feat}} \) is set to 40. Finally, another \( conv \) \( 1\times1 \) is used for downsampling to recover the \( 3 \)-channel illumination mapping \(\bar L\), which is then element-wise multiplied by the low-light image \( I \) to obtain the illuminated image \( I_{\text{lu}} \).

\subsubsection{Illumination-Fused Vison Mamba.}
The structure of the Damage Restorer (IFVM) is shown in Fig. \ref{fig:enter-label}(b), which consists of an encoder and a decoder based on the illumination fusion visual mamba. The encoder represents the downsampling process, while the decoder represents the upsampling process. The upsampling and downsampling processes are symmetric and divided into two levels. Firstly, the illuminated image \( I_{\text{lu}} \) obtained from the Illumination Estimator IE is downsampled by a \( conv \) \( 3\times3 \) (stride = 2) to match the dimension of the illumination feature map \( F_{\text{lu}} \), facilitating subsequent operations. Next, we perform downsampling to reduce computational complexity and extract deep features. The downsampling process is divided into two levels, each level consisting of an Illumination Fusion State Space Model (IFSSM) and a convolutional layer with a stride of 2 and kernel size of \( 4 \times 4 \). After each convolutional layer, the width and height of the image are halved, while the feature dimension is doubled. Therefore, after two levels of downsampling, the deepest feature dimension should be \( 4C \). After extracting image features, we need to perform upsampling to recover the image. Similar to downsampling, upsampling is also divided into two levels, each consisting of a \( deconv \) \( 2\times2 \) (stride = 2) and a \( conv \) \( 1\times1 \), and an Illumination Fusion State Space Model (IFSSM). After each deconvolution layer, the width and height of the image are doubled, while the feature dimension is halved. The output of the deconvolution layers is then linked with the corresponding level of downsampling Illumination Fusion State Space Model (IFSSM) output to mitigate the loss of image information during downsampling. Finally, a \( conv \) \( 3\times3 \)(stride = 2) is applied to the image to reduce the dimensionality and restore it to the RGB format with three channels. The enhanced image \( I_{\text{en}} \) is obtained by performing a residual connection with the restored image and \(I_{\text{lu}}\).

\subsection{Illumination-Fused State Space Models.}
\begin{figure}
    \centering
    \includegraphics[width=0.71\linewidth]{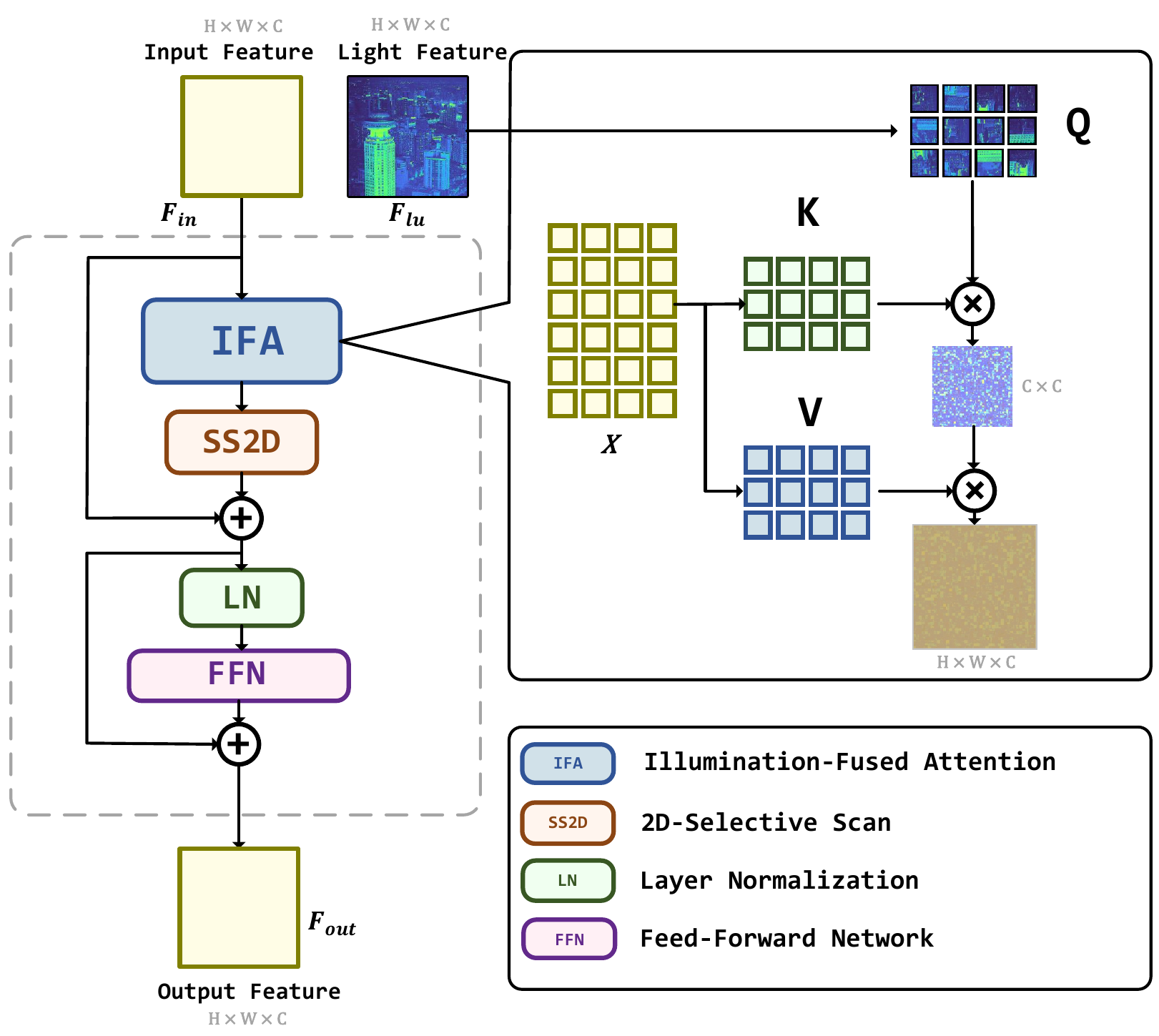}
    \caption{Our Illumination Fusion State Space Model (IFSSM) integrates lighting features and input vector $x$ using a fused-block, and utilizes the linear 2D Selective Scanning model (SS2D) for feature extraction. In IFA, we treat lighting features as $Q$, and input vectors as $KV$ to calculate attention scores.
}
    \label{fig:fig.2}
\end{figure}

\hspace{1em}In low-light enhancement research, Convolutional Neural Networks (CNNs) have limitations in processing overall image information. Transformers may affect the efficiency of practical applications due to their high computational demands. To address this, our designed Illumination Fusion State Space Model (IFSSM) includes Fused Attention (IFA), an SS2D module, an LN layer, a Feed-Forward Network (FFN), and a convolutional layer to match the dimensions of illumination feature maps with inputs, as shown in Fig. \ref{fig:fig.2}.

\subsubsection{Illumination-Fused Attention.} 
As shown in Fig.\ref{fig:fig.2}, the illumination feature map \( F_{\text{lu}} \) generated by IE is input together with the brightened image after feature extraction into IFA. In \cite{ref36}, the illumination feature map \( F_{\text{lu}} \) is used as a token, transformed and multiplied with \( V \)(value) generated from \( x \) to calculate attention scores. However, this method results in \( KV \)(key and value) not coming from the same input, violating the principle in Transformers of ensuring that all information processing focuses on the content of the input data rather than external or irrelevant information to the current task, breaking data consistency and alignment, and the additional multiplication operation also increases parameter requirements and computational complexity. Therefore, our IFA adopts the Cross Attention mechanism, first adjusting \( F_{\text{lu}} \) and input \( x \) to a form suitable for multi-head attention by changing dimensions:
\begin{equation}
\mathbf{X} = [X_1, X_2, \ldots, X_k], \quad \mathbf{F}_{lu} = [F_{lu1}, F_{lu2}, \ldots, F_{luk}]
\end{equation}
where \(X_i \in \mathbb{R}^{HW \times d_k}\),\(F_{lui} \in \mathbb{R}^{HW \times d_k}\),\( k \) is the number of heads, \( d_k \) denotes the dimensionality of each head, and \( d_k \) = \(\frac{C}{k}\), with \( C \) representing the dimensionality of the input features.We note that using an illumination prior map as the query vector allows the model to more specifically address the dark areas in the image. Consequently, the attention mechanism of the model can focus on the low-light areas that need enhancement, rather than processing the entire image uniformly. Therefore, we treat \( F_{\text{lu}} \) as \( Q \in \mathbb{R}^{HW \times d_k}\) and the input \( x \) as \( K,V \in \mathbb{R}^{HW \times d_k}\), to fuse the two input features, enabling \( F_{\text{lu}} \) to guide the self-attention calculation of \( x \).

\begin{equation}
Q_i = \mathbf{F}_{lui} W_{Q_i}, \quad K_i = X_i W_{K_i}, \quad V_i = X_i W_{V_i},
\end{equation}
where \( W_Q \), \( W_K \), and \( W_V \in \mathbb{R}^{d_k \times d_k}\)  are the learnable parameter matrices constructed by convolutional layers. Thus, the self-attention of each head can be represented by the formula:

\begin{equation}
\text{Attention}(Q_i, K_i, V_i) = (V_i)\text{softmax}\left(\frac{K_i^T Q_i}{\alpha_i}\right),
\end{equation}
where \( \alpha_i \in \mathbb{R}^{1}\) is a learnable parameter serving as a scaling factor to adjust the attention scores, thereby controlling the sharpness of the attention weights, the \( k \) heads are subsequently reshaped back to the standard image format (\( B, C, H, W \)) and aggregated via a convolutional layer, resulting in an output with dimensions that match the original.

\subsubsection{2D-Selective Scan.}
Inspired by the SSM model in \cite{ref40}, we adopted the approach of integrating an SSM into visual tasks from \cite{ref34}. The SS2D module in \cite{ref34} consists of Scan Expanding, S6 blocks, and Scan Merging operations. For the processed images we input, they first undergo the Scan Expanding operation, where the image is unfolded from its four corners as shown in Fig. \ref{fig:fig.5}. The images are then flattened, meaning the height (H) and width (W) are merged into a token length (L). Each sequence from the scan is then input into an S6 module for feature extraction. The calculation formula for S6 can be expressed as:
\begin{align}
h_t &= \bar{A} h_{t-1} + \bar{B} x_t \\
y_t &= \bar{C} h_t
\end{align}
where \( x \) is the input variable, \( y \) is the output, and \( \bar{A} \),\( \bar{B} \) and \( \bar{C} \) are all learnable parameters.Afterward, the outputs from the four directions of extracted features \(y_1, y_2, y_3, y_4\) are summed and merged, and the dimensions of the merged output are readjusted to match the input size. Furthermore, to extract deeper latent features, we set the number of hidden layers in SS2D to increase with each level of IFSSM. We default the number of hidden layers \(d_{\text{state}}\) in SS2D to 16, doubling the number of layers with each sampling level, thus reaching up to 64 layers at the deepest sampling. This setting allows for progressively deeper feature extraction from the vector that has integrated illumination features.

\section{Experiment}
\subsection{Datasets and Implementation details}
\begin{figure}
    \centering
    \includegraphics[width=1\linewidth]{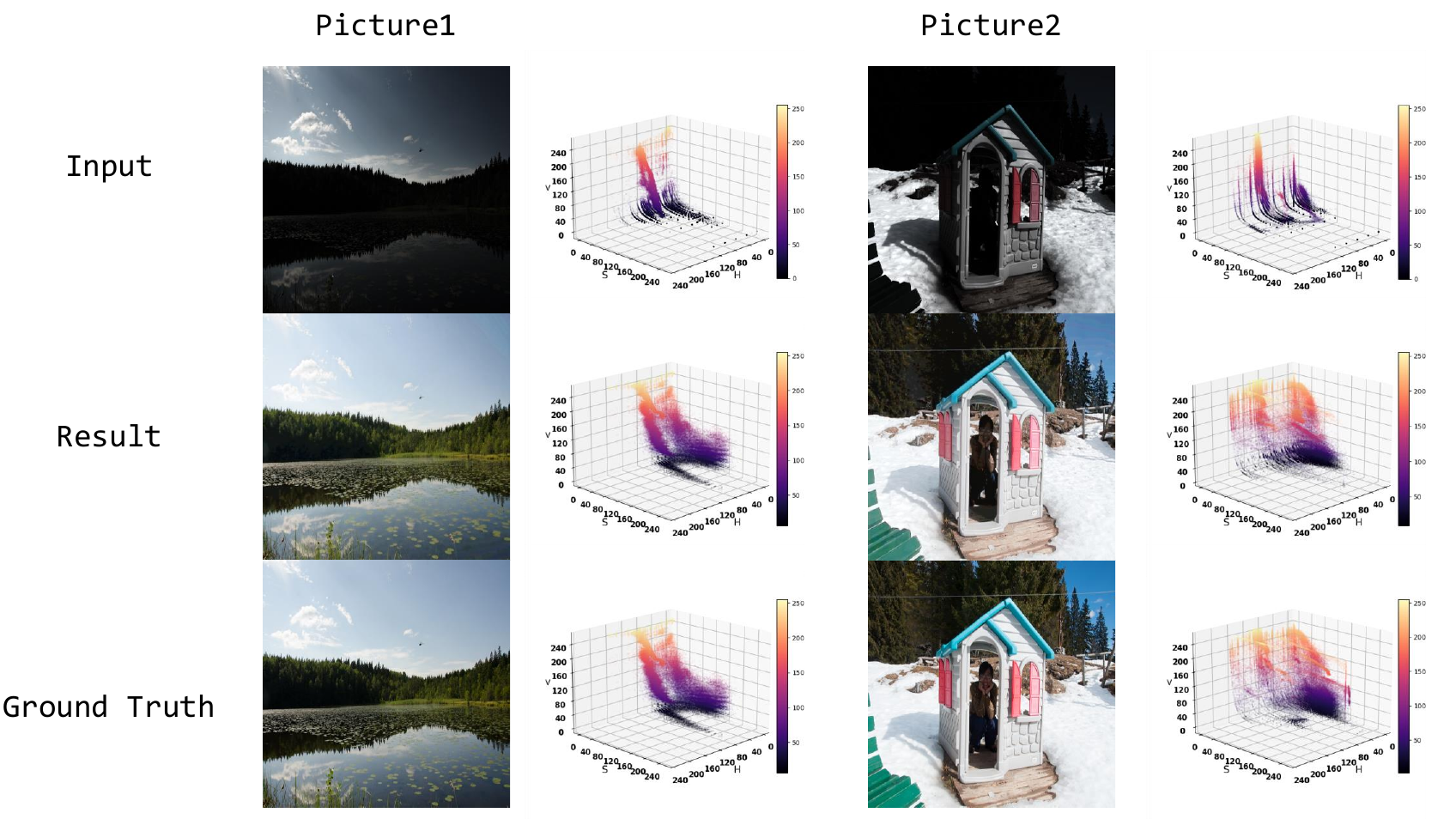}
    \caption{3D image display of the HSV color space.}
    \label{fig:fig.7}
\end{figure}
\hspace{1em}We evaluated the performance of our model on the LOL dataset (v1\cite{ref30} and v2\cite{ref52}).
\subsubsection{LOL.}
The LOL dataset is divided into two versions, v1 and v2. In LOL\_v1, the ratio of training to test data pairs is 485:15. Each pair consists of an input low-light image and a target reference image. The LOL\_v2 dataset is further divided into LOLv2\_real and LOLv2\_synthetic. The ratios of training to test data pairs in LOLv2\_real and LOLv2\_synthetic are 689:100 and 900:100, respectively. The distribution of data pairs is the same as in LOL\_v1.

\subsubsection{Implementation Details.}
We implemented our RetinexMamba model in Pytorch and trained and tested it on a PC with A10 and V100 GPUs under a Linux system (CUDA 11.7, Python 3.8, Pytorch 1.13). We set the resolution of images to 128x128. The batch sizes for LOL\_v1 and LOLv2-synthetic were set to 8, and for LOLv2\_real to 4. Standard augmentation methods such as random rotation and flipping were used to enhance the training data. To minimize loss, we employed the Adam optimizer with a momentum term \(\beta_1\) set at 0.9 and RMSprop control parameter \(\beta_2\) set at 0.999, aiming to minimize the Mean Absolute Error (MAE) between the enhanced images and the ground truth. Additionally, a cosine annealing schedule was used to prevent the loss from getting stuck in local minima.

\begin{table}[ht]
\centering
\caption{Quantitative comparisons on LOL-v1 and LOLv2-real}
\label{tab:mytable1}
\setlength{\tabcolsep}{12pt} 
\resizebox{\textwidth}{!}{%
\begin{tabular}{c|c c c|c c c}
\hline
\multirow{2}{*}{\textbf{Methods}} & \multicolumn{3}{c|}{\textbf{LOL-v1}} & \multicolumn{3}{c}{\textbf{LOLv2-real}} \\
 & \text{PSNR} $\uparrow$ & \text{SSIM} $\uparrow$ & \text{RMSE} $\downarrow$ & \text{PSNR} $\uparrow$ & \text{SSIM} $\uparrow$ & \text{RMSE} $\downarrow$ \\
\hline
\textbf{Supervised} & & & & & & \\
LIME\cite{ref17} & 16.362 & 0.624 & 21.07 & 16.342 & 0.653 & 22.54 \\
MBLLEN\cite{ref21} & 17.938 & 0.699 & 18.78 & 15.950 & 0.701 & 30.22 \\
Retntinex-Net\cite{ref22} & 17.188 & 0.589 & 22.59 & 16.410 & 0.640 & 20.21 \\
KinD\cite{ref23} & 20.347 & 0.813 & 14.30 & 18.070 & 0.781 & 18.04 \\
KinD++\cite{ref30} & 20.707& 0.791 & 14.34 & 16.800 & 0.741 & 15.64 \\
MIRNet \cite{ref26} & 24.140 & 0.842 & 12.03 & 20.357 & 0.782 & 14.21 \\
URetntinex-Net\cite{ref33} & 21.450 & 0.795 & 13.55 & 21.554 & 0.801 & 14.23 \\
Retinexformer\cite{ref36} & 23.932 & \textcolor{red}{0.831}  &  8.35 & 21.230 & 0.838  &  9.92 \\
\textcolor{red}{RetinexMamba} & \textcolor{red}{24.025} & 0.827  & \textcolor{red}{8.17} & \textcolor{red}{22.453} & \textcolor{red}{0.844} & \textcolor{red}{9.38}\\
\hline
\textbf{Unsupervised} & & & & & & \\
GenerativatePrior \cite{ref60} & 12.552 & 0.410 & 45.80 & 13.041 & 0.552 & 40.23 \\
Zero-Dce \cite{ref27} & 16.760 & 0.560 & 34.42 & 18.059 & 0.580 & 29.01 \\
RUAS \cite{ref32} & 16.401 & 0.503 & 30.21 & 16.873 & 0.513 & 29.23 \\
SCI \cite{ref37} & 14.864 & 0.542 & 24.87 & 15.342 & 0.521 & 27.50 \\
PairLie \cite{ref58} & 19.691 & 0.712 & 19.03 & 19.288 & 0.684 & 20.01\\
NeRCO \cite{ref57} & 19.701 & 0.771 & 24.80 & 19.234 & 0.671 & 23.13 \\
CLIP-LIE \cite{ref59} & 17.207 & 0.589 & 10.18 & 17.057 & 0.589 & 10.64 \\
Enlighten-Your-Voice  \cite{ref69} &19.728 & 0.715 &10.13& 19.335 &0.686 & 10.21 \\
\hline
\end{tabular}
}
\end{table}

\begin{figure}
    \centering
    \includegraphics[width=1\linewidth]{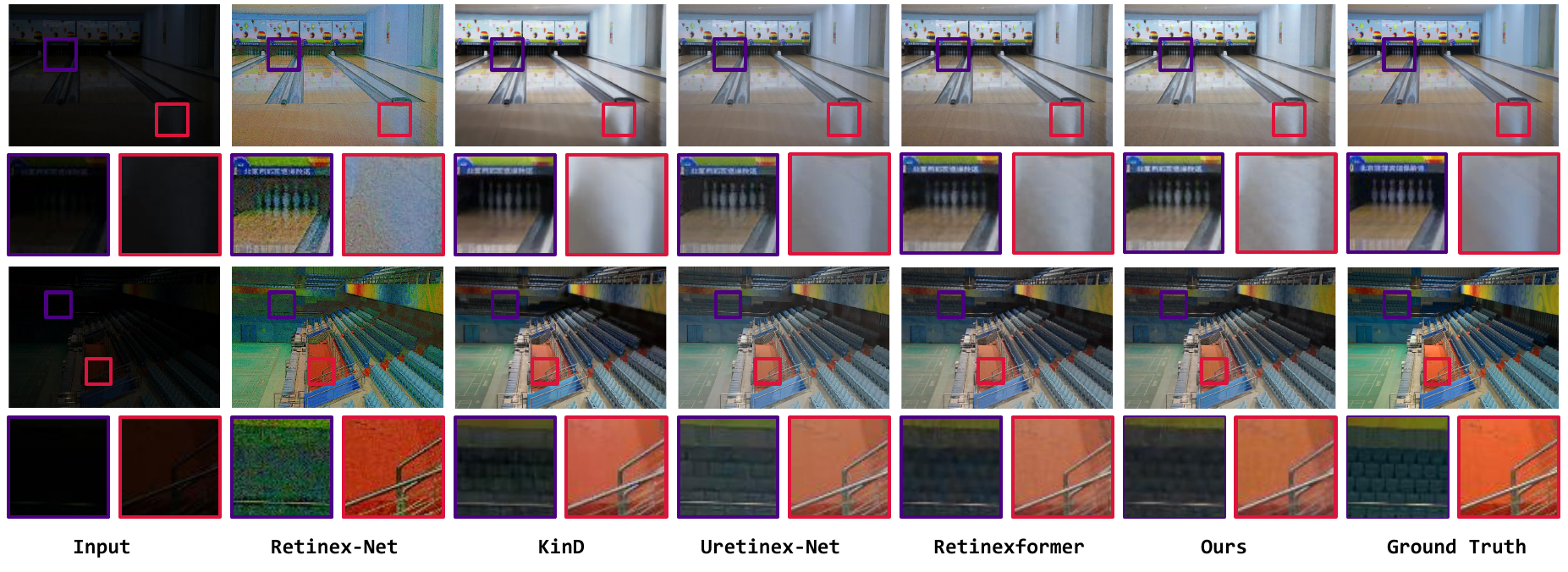}
    \caption{The above are the qualitative experimental results on LOLv1. Our method effectively reduced color distortion and enhanced the lighting effects.
}
    \label{fig:fig.3}
\end{figure}

\subsection{Low-light Image Enhancement}
\hspace{1em}We compare our method against various SOTA methods in both supervised and unsupervised domains in Tab.\ref{tab:mytable1}.The datasets used for comparison are the synthetic data from LOLv1 and the real data from LOLv2. All data in the table were obtained under the same conditions using publicly available code for training and testing or derived from the original papers. The results indicate that our method outperforms the aforementioned SOTAs in terms of PSNR and RMSE, while SSIM is slightly lower than that of Retinexformer.

\subsubsection{Quantitative Results.}
The metrics we used for comparison include PSNR, SSIM, and RMSE. A higher PSNR indicates better image enhancement effects, while a higher SSIM indicates the preservation of more high-frequency details and structures in the results. A lower RMSE value signifies better performance of the prediction model, as it indicates smaller errors. Compared to the baseline and the best existing technology method\cite{ref36}, our method achieved an increase in PSNR by 0.093 and 0.77 on the LOL\_v1 and LOLv2\_real datasets, respectively. RMSE decreased by 0.39 on the LOLv2\_real dataset, which is desirable as lower values are better.

\begin{figure}
    \centering
    \includegraphics[width=1\linewidth]{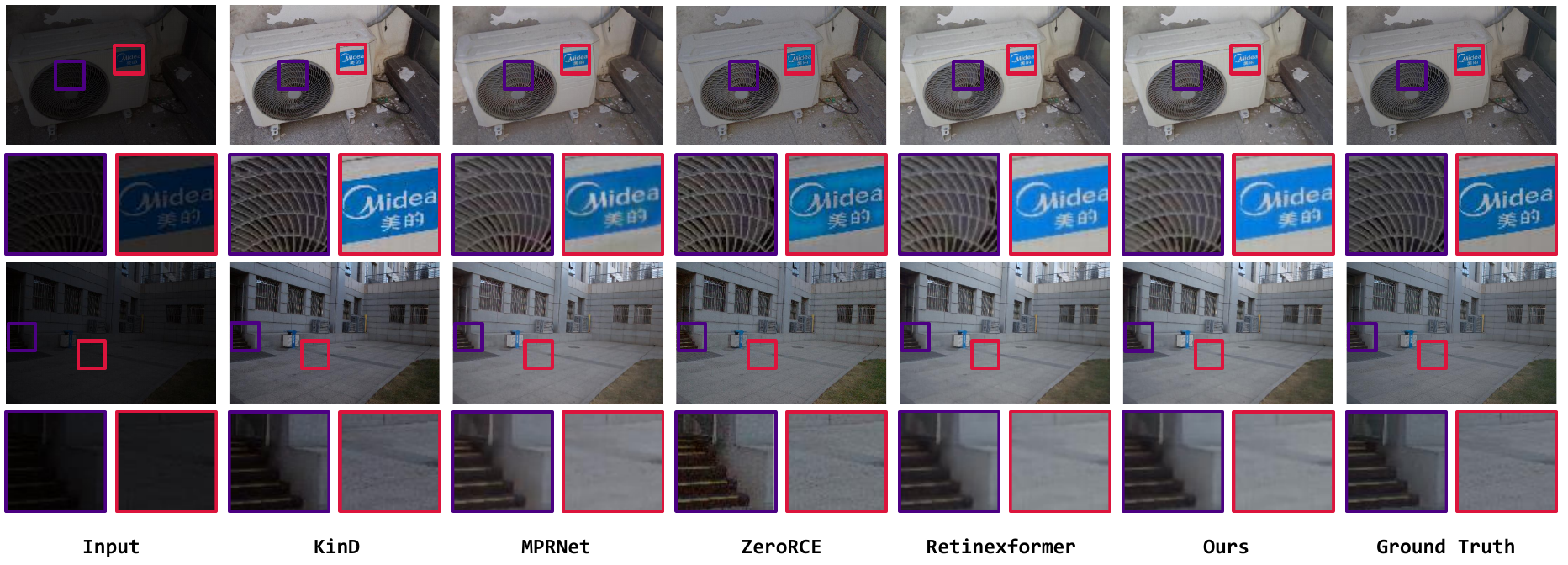}
    \caption{The above are the qualitative experimental results on LOLv1. Our method effectively reduced color distortion and enhanced the lighting effects.
}
    \label{fig:fig.4}
\end{figure}
\subsubsection{Qualitative Results.}
\begin{figure}
    \centering
    \includegraphics[width=1\linewidth]{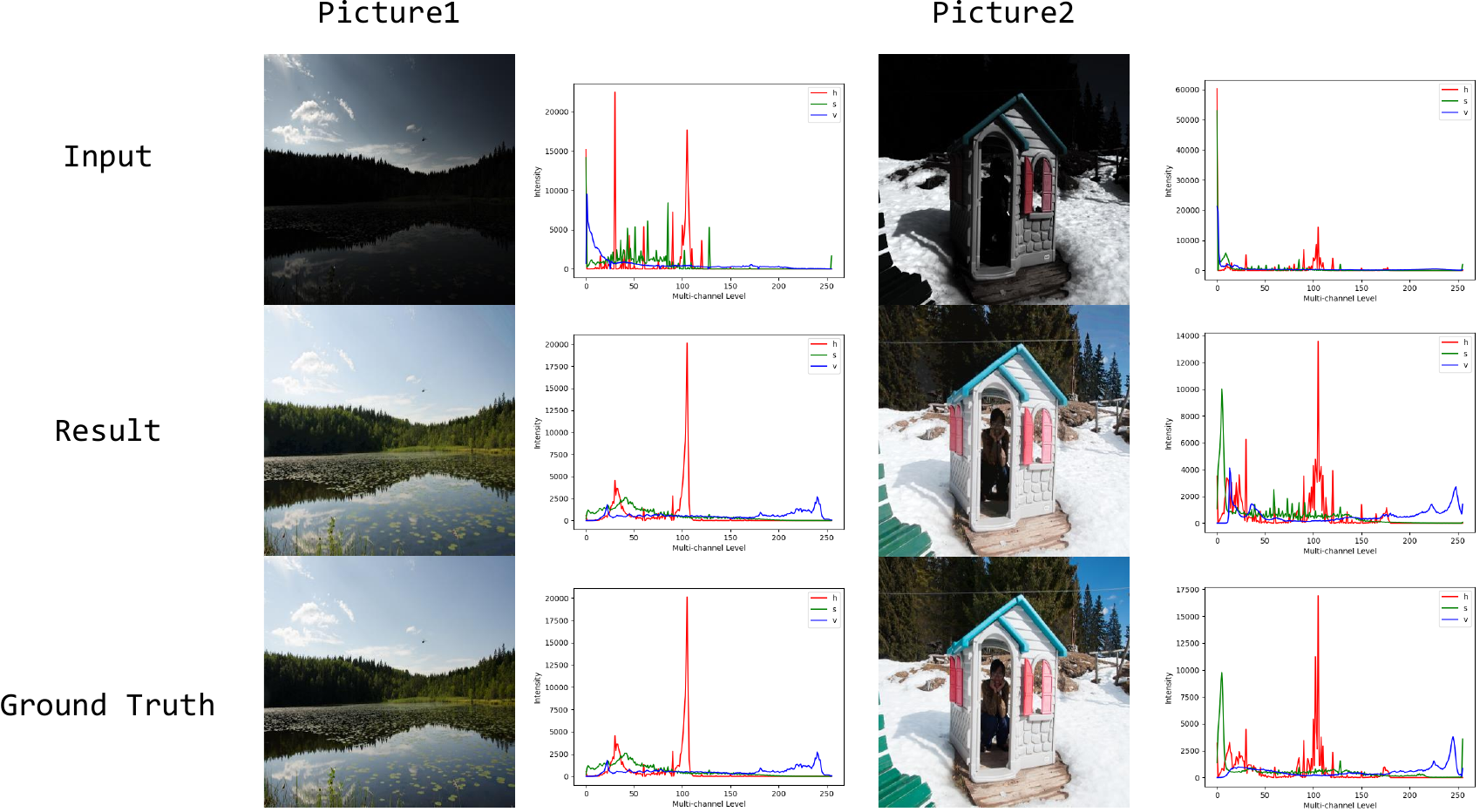}
    \caption{2D image display of the HSV color space.}
    \label{fig:fig.6}
\end{figure}

The qualitative comparison results between RetinexMamba and other SOTA algorithms are shown in Fig. \ref{fig:fig.3} and \ref{fig:fig.4}. Please zoom in for better visual clarity. Figure Three compares the LOLv1 dataset, while Fig. \ref{fig:fig.4} compares the LOLv2\_real dataset. As depicted in Figure Three, previous methods exhibit amplification of noise, such as in Retinex-Net, as well as cases of underexposure like KinD and overexposure like Uretinex-Net. Similarly, on the synthetic dataset in Fig. \ref{fig:fig.4}, KinD shows instances of overexposure in the air conditioner area at the top and underexposure in the staircase area at the bottom, while ZeroRCE exhibits extensive noise and artifacts. Likewise, in Retinexformer, underexposure around the bowling ball at the top of Fig. \ref{fig:fig.3} and color distortion in the stadium at the bottom are observed. In contrast, our RetinexMamba effectively controls exposure intensity, reduces color distortion, and minimizes noise. Additionally, we compared two images on the LOLv2\_syn dataset in both 2D and 3D HSV color space in Figs. \ref{fig:fig.6} and \ref{fig:fig.7}. The results demonstrate that the images enhanced by RetinexMamba are closest to the Ground Truth.

\subsection{Ablation Study}
\begin{table}[ht]
\centering
\caption{Ablation Study on LOL-v1,LOLv2-real and LOLv2-syn}
\setlength{\tabcolsep}{3pt} 
\resizebox{\textwidth}{!}{%
\begin{tabular}{c|c c c|c c c|c c c}
\hline
\multirow{2}{*}{\textbf{Methods}} & \multicolumn{3}{c|}{\textbf{LOL-v1}} & \multicolumn{3}{c|}{\textbf{LOLv2-real}} & \multicolumn{3}{c}{\textbf{LOLv2-syn}} \\
 & \text{PSNR} $\uparrow$ & \text{SSIM} $\uparrow$ & \text{RMSE} $\downarrow$ & \text{PSNR} $\uparrow$ & \text{SSIM} $\uparrow$ & \text{RMSE} $\downarrow$ & \text{PSNR} $\uparrow$ & \text{SSIM} $\uparrow$ & \text{RMSE} $\downarrow$ \\
\hline
Ours FixedHS & 23.839 & 0.823  & 8.29  &  21.756  &  0.832 & 9.51  & 25.419  & 0.928  & 8.34\\
Ours NoFB & 22.730 &  0.826 & 8.70 & 22.262  & 0.823  & \textcolor{red}{9.27} & 25.343 & 0.934 & 8.40 \\
Ours NoSS2D & 23.035 &  0.821 & 8.75 & 21.309 & 0.807  & 9.74 & 24.861  & 0.923  & 8.43 \\
Ours IGMSA & 22.054 &  0.797 & 9.51 & 21.745  & 0.824  & 9.46 & 25.236  & 0.928 & 8.43 \\
\textcolor{red}{RetinexMamba} & \textcolor{red}{24.025} & \textcolor{red}{0.827} & \textcolor{red}{8.17}& \textcolor{red}{22.453} & \textcolor{red}{0.844} & 9.38 & \textcolor{red}{25.887} & \textcolor{red}{0.935} & \textcolor{red}{8.24} \\
\hline
\end{tabular}
}
\end{table}
\hspace{1em}We conduct our ablation studies on three datasets: LOLv1, LOLv2\_real, and LOLv2\_syn. We have set up six different architectures here, each varying by removing different components or setting different hyperparameters for comparison.

\begin{itemize}
    \item "Ours FixedHS" denotes that we have set the number of hidden layers \(d\_state\) in the SS2D module of IFSSM to a fixed value of 16, which no longer deepens as the sampling levels increase.
    \item "Ours NoFB" indicates that we have removed the fused-block in IFSSM and used element-wise multiplication to fuse the illumination features and the input vector.
    \item "Ours NoSS2D" indicates that we have removed the SS2D module from IFSSM, retaining only the Transformer architecture.
    \item "Ours IGMSA" denotes that we have replaced the fused attention in IFSSM with the Illumination-Guided Multi-head Self-Attention (IG-MSA) from \cite{ref36}.
\end{itemize}
\hspace{1em}Compared to all ablation results, our ablation setup achieved the highest PSNR and SSIM. "Ours FixedHS" demonstrated the drawbacks of insufficient feature extraction and the inability to capture long sequences with a fixed number of hidden layers in the SS2D model. "Ours NoFB", which uses direct element-wise multiplication to fuse illumination features, lacks a logical explanation. Meanwhile, "Ours NoSS2D" and "Ours IGMSA" each highlight the disadvantages of using only the Transformer architecture and the poor interpretability of attention calculations from \cite{ref36}, respectively.

\section{Conclusion}
\hspace{1em}In this paper, we introduce the RetinexMamba architecture, which is based on the Retinexformer and Mamba to enhance low-light images. Initially, building on the Retinexformer model, we divided it into an illumination estimator and a damage restorer, and inspired by VMamba, we incorporated the SS2D model to address the inherent position sensitivity of visual data in Transformers. Furthermore, we replaced the IG-MSA module with a more interpretable Fused-Attention to fuse illumination features and the input vector. Extensive quantitative and qualitative experiments demonstrate that our RetinexMamba outperforms the current state-of-the-art on the LOL dataset. Although the computational complexity of SS2D is reduced, the overall number of parameters has increased, consuming more computational resources. Therefore, our future work will focus on reducing the total number of parameters while maintaining computational complexity.
%
%
\bibliographystyle{splncs04}
\bibliography{reference.bib}
\end{document}